\documentclass[11pt,a4paper]{article}
\usepackage[hyperref]{eacl2021}
\usepackage{times}
\usepackage{latexsym}

\usepackage{microtype}

\usepackage{multirow}
\usepackage{graphicx}
\usepackage{enumitem}

\aclfinalcopy % Uncomment this line for the final submission
\title{SentEmojiBot: Empathising Conversations Generation with Emojis}

\author{Akhilesh Ravi, Amit Yadav, Jainish Chauhan, Jatin Dholakia, Naman Jain, and Mayank Singh \\
  Indian Institute of Technology Gandhinagar \\
  Gujarat, India \\
  \texttt{akhilesh.ravi@iitgn.ac.in} \\}

\date{}

\begin{document}
\maketitle
\begin{abstract}
The increasing use of dialogue agents makes it extremely desirable for them to understand and acknowledge the implied emotions to respond like humans with empathy. Chatbots using traditional techniques analyze emotions based on the context and meaning of the text and lack the understanding of emotions expressed through face. Emojis representing facial expressions presents a promising way to express emotions. However, none of the AI systems utilises emojis for empathetic conversation generation. We propose, SentEmojiBot, based on SentEmoji dataset, to generate empathetic conversations with a combination of emojis and text. Evaluation metrics show that BERT-based model outperforms the vanilla transformer model. A user study indicates that the dialogues generated by our model were understandable and adding emojis improved empathetic traits in conversations by 9.8\%.
\end{abstract}

\section{Introduction}

Humans acknowledge the feelings of their interlocutor while responding with caring attitude to achieve an engaging and comforting conversation. This behaviour is termed as empathetic responding \cite{Rashkin2018}. With the onset of technologies such as chatbots and voice assistants, humans have started to expect empathetic responses from the machine-mediated automatic communication systems \cite{Reeves1996}. Many studies have proved that empathetic responses results in better outcomes from both goal-oriented and informal conversations. \cite{Levinson2000,wentzel1997student,bickmore2001relational,kim2004effects,fraser2018spoken}. In recent years, researchers have been successful in generating meaningful responses \cite{Zhou2018,Wang2018,Zhou2018a,Hu2017} and embedding empathetic behaviour in the semantics of a chatbot's response \cite{Ritter2010,Zhang2018,Mazare2019,Rashkin2018,Lin2019}. However, these works have been able to generate responses by focusing purely on textual responses.

% been successful in generating meaningful empathetic responses. However, they are still limited to utilising textual data, which limits in expressing emotions and setting response tone in a chat-based environment. 
\begin{figure}[!b]
\centering
\resizebox{\hsize}{!}{
\includegraphics{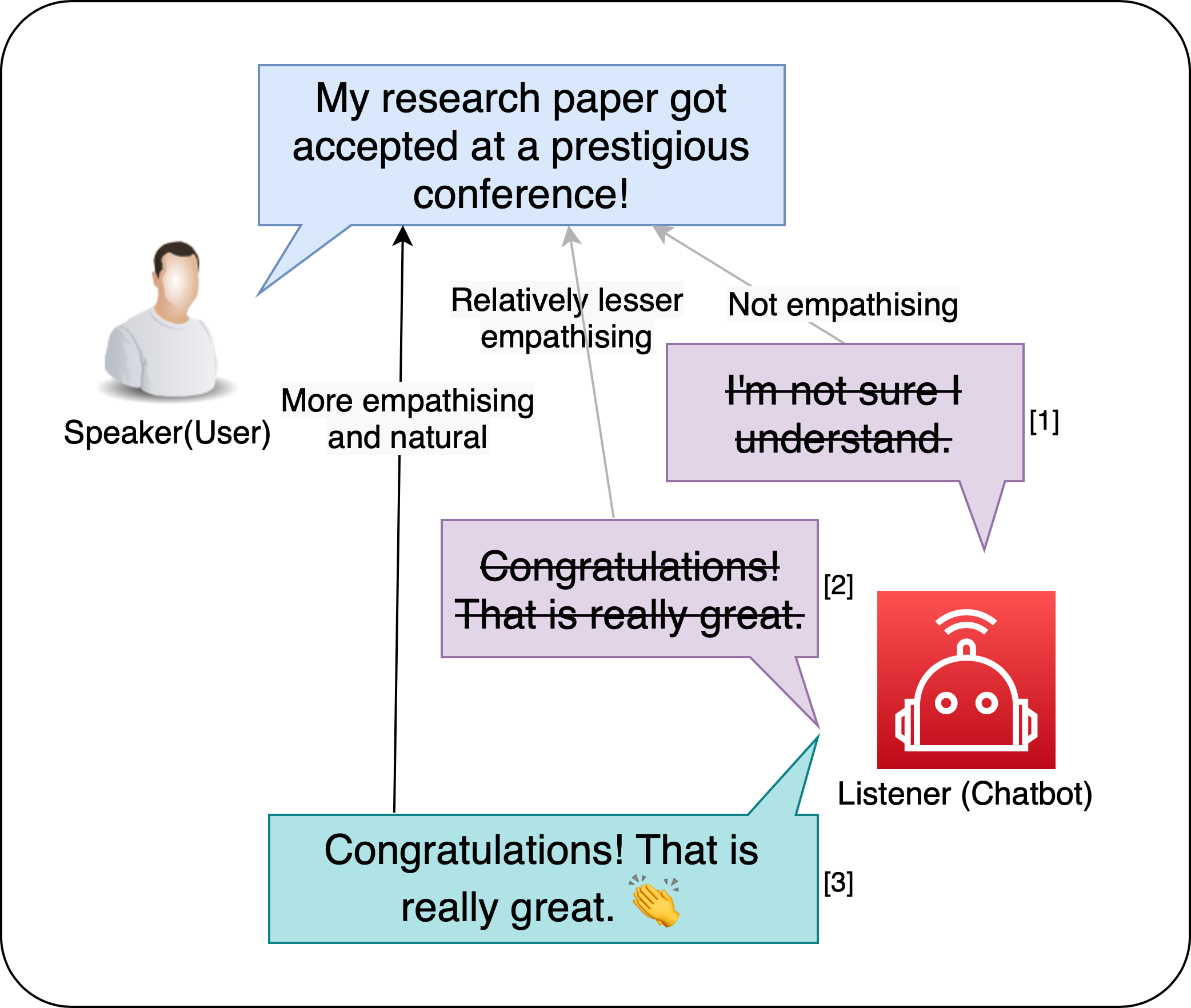}}

\caption{Comparison of responses from various systems: 1) Siri, 2) \citet{Rashkin2018}, 3) Our model} 
\label{fig1}
\end{figure}

Research shows that facial expressions plays a key role in clearly communicating the message of the speaker \cite{10.1145/1027933.1027968}. They help the listener to clearly resolve the ambiguity in emotions, intention and tonality of the message. Modern application softwares have introduced Emojis, the animated faces with expressions, as an alternative to facial expressions in chat rooms to eliminate the ambiguity related to the response of the user. Previous works have analysed and supported the significance of emojis in social media conversations through improved performances in understanding NLP tasks such as sentiment, emotion, and sarcasm detection \cite{Felbo_2017, wood2016emoji,li2019exploring}. Even though we find rich literature that use emojis to improvise semantic understanding of text, to the best of our knowledge, we did not find any work that uses emojis to enhance the generation of empathetic responses in automated communication systems.

In this paper, we formalise the task of generating empathising responses using emojis by proposing \textit{SentEmojiBot}, a model trained on textual conversations and emojis data. We present the experiments with appropriate evaluation methods to prove the significance of emojis in conveying empathising messages. Figure~\ref{fig1} shows an example of a chatbot interface where Speaker(human) initiates the conversation. The figure compares various systems and clearly shows the positive impact of empathising text and emojis through the gradual improvement in empathetic behaviour from Siri to \textit{SentEmojiBot}. \textit{SentEmojiBot} is a BERT-based model that generates responses based on the emotion and context of the text. In our experiments, the BERT based model outperformed the vanilla transformer model. Moreover, a user survey shows that \textit{SentEmojiBot} added relevant emojis to conversations which improved the empathising behaviour of the responses by 9.8\%, compared to purely text-based response. Hence, our work showcases the possibility of building natural, engaging, and empathetic dialogue agents over the traditional text-based language models.

Our main contributions are SentEmojiBot - a pipeline for generating empathetic responses with emojis, and a user-study showing an increase in empathetic behaviour when emoji is added to a textual traditional response.

\begin{figure}[!t]
\centering
\includegraphics[width=0.95\columnwidth]{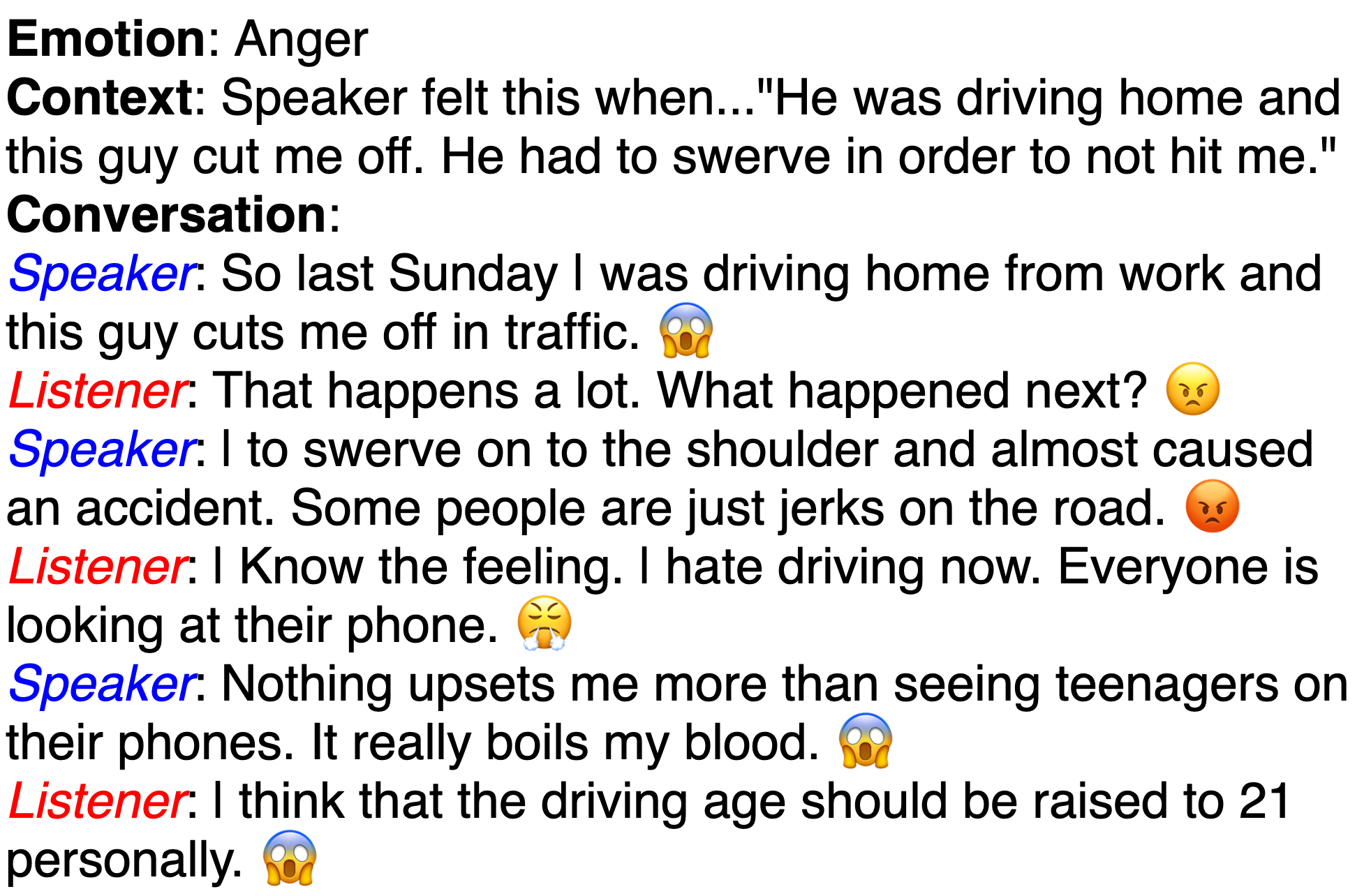}
\caption{Example of a conversation snippet with multiple utterances from SE dataset}
\label{fig2}
\vspace{-0.5cm}
\end{figure}

% The code is available at \url{https://github.com/JatinDholakia/SentEmojiBot}. 
%Embedded Links are not allowed. Thus, we are not using hyperlinks.
%Links are not allowed during review process.

\section{Dataset}

We utilise SentEmoji (hereafter \textit{`SE'}) dataset released by \citet{10.1145/3371158.3371218} containing empathetic responses with emojis. The dataset contains 24,850 conversations and 79,190 utterances, with an average utterance length of  15.2 words. The dataset has 10 fundamental emotional categories. These categories are mutually exclusive from each other, in terms of appraisal, antecedent events, probable behavioural response and physiology \cite{Kowalska2017}.  
Figure~\ref{fig2} presents an example of conversation snippet from the SE dataset. ``Emotion'' tells about the implied emotion in the conversation. ``Context'' sets a situation for conversation based on the emotion. In every conversation, ``Speaker'' refers to human and ``Listener'' refers to automated dialogue agent. Each dialogue is considered as one utterance and each utterance contains an emoji to either highlight the speaker's emotion or generate empathetic response from the listener.

\section{Methodology}
This section discusses the experimental setup and the architecture of SentEmojiBot (Figure \ref{fig:fig5}). 

% We approach the experiments by dividing the task into two subtasks - a) generating a textual response and b) embedding the relevant emoji to the response. To generate a text response, we show experiments with BERT-based architecture \cite{Wolf2019HuggingFacesTS} and a Transformer-based retrieval architecture \cite{Vaswani2017}. For adding emoji, we need to identify the relevant emotion in the text. We use a CNN-based classifier to obtain the emotion. We use the generated text from the language model as an input and the "Emotion" from each conversation snippet as the output to train the classifier. Next, we obtain the emoji bucket corresponding to the detected emotion of the response based on Table \ref{table2}\todo{Where does the the table comes from? SE dataset or you made it in this paper? or any other paper?}. Finally, we obtain the most appropriate emoji using the cosine similarity between each emoji's embedding \cite{Eisner2016} and average pooled text embedding \cite{Demeester2016}. \todo{Are you using Bert or word2vec?} Figure \ref{fig5} depicts the architecture of SentEmojiBot. 

% We discuss further in detail about the preprocessing, architecture description, and the training settings.

\subsection{Data Preparation}
% Figure \ref{fig2} shows a conversation snippet from SE dataset with 3 Speaker and 3 Listener utterances.
In a conversation, people only have the information about the utterances, with their interlocutor, that have been discussed in the past in order to analyse and convey their response in return. Hence, we concatenate utterances prior to the listener's response, from the SE's conversations as the ``context utterance'' and the listener's response as the ``response utterance''. The context utterance is fed as an input to the model to obtain response utterance as an output. In total, there are 53,372 context-response utterance pairs. We do not use emotion and context in the training process and do not consider speaker's response as the ``response utterance'' because speaker drives the conversation for the listener and expects a response in return. Also, in the real world deployment of SentEmojiBot, listener is expected to be an automated model output whereas speaker is expected to be a human. We tokenised the context utterance using the BertTokenizer \cite{Wolf2019HuggingFacesTS} and the sequence length is set to 100. The result is fed to the language models described below to get an empathetic response.

%  To tokenize the utterances, we used BertTokenizer \cite{Wolf2019HuggingFacesTS} which uses WordPiece embeddings with a 30,000\todo{write the exact number} token vocabulary. 

\subsection{Generating ``Response Utterance''}
To generate an empathetic text response, we perform experiments on retrieval-based systems consisting of Transformers. In retrieval-based systems, the model selects the best possible response from a set of candidate responses. The following methodology has been formalised by \citet{Rashkin2018}.

\begin{figure*}[!t]
\begin{center}
\resizebox{0.95\hsize}{!}{
\includegraphics{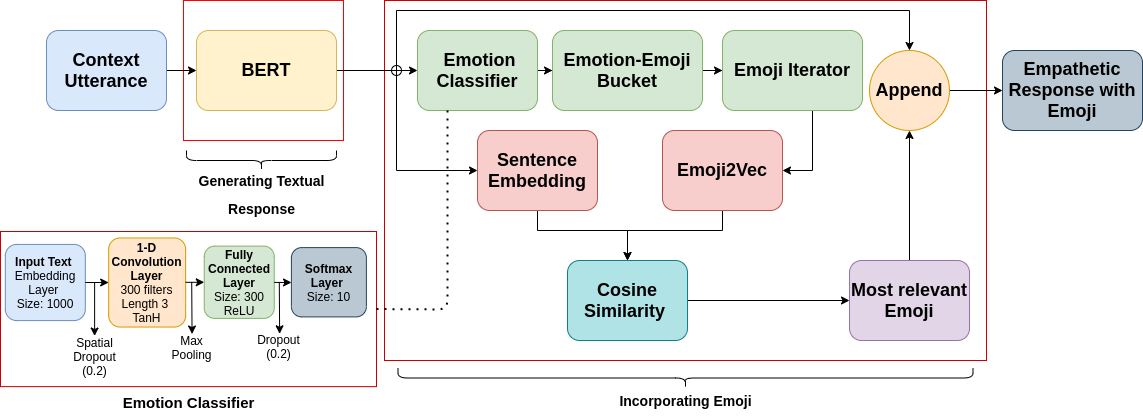}}
\caption{Architecture of SentEmojiBot}
\label{fig:fig5}
\end{center}
\vspace{-0.5cm}
\end{figure*}

\begin{itemize}
    \item \textbf{BERT-based}: We used BERT \cite{Devlin2018} as the base architecture to encode candidates (h$_y$) and contexts (h$_x$). The model is fine-tuned over pre-trained weights \cite{Wolf2019HuggingFacesTS} on SE dataset, all layers are trained for 12 epochs with a batch size of 16, an embedding layer of size 300, the learning rate of $5 \times 10^{-5}$, and the Adamax optimizer.
    
    \item \textbf{Vanilla Transformers-based}: We use two transformer encoders separately embedding context (h$_x$) and candidates (h$_y$) \cite{yang2018learning}. The learning rate is set to $8 \times 10^{-4}$, with an Adamax optimizer. The model is fine-tuned for 25 epochs with a batch size of 128.
    
\end{itemize}
We provide the ``context utterance'' as an input and predict the next most probable ``response utterance'' from the model. The model chooses a response according to a softmax on the dot product (h$_x$·h$_y$) out of all candidates. We minimise the negative log-likelihood of selecting the correct response. The utterances from the SE dataset were split into three parts: training data (80\%), validation data (10\%) and test data (10\%). The number of training epochs was decided to avoid over-fitting on the data and due to resource constraints.

\begin{itemize}
    \item \textbf{BERT-based}: We used BERT \cite{Devlin2018} as the base architecture to encode candidates (h$_y$) and contexts (h$_x$). The model is fine-tuned over pre-trained weights \cite{Wolf2019HuggingFacesTS} on SE dataset, all layers are trained for 12 epochs with a batch size of 16, an embedding layer of size 300, the learning rate of $5 \times 10^{-5}$, and the Adamax optimizer.
    
    \item \textbf{Vanilla Transformers-based}: We use two transformer encoders separately embedding context (h$_x$) and candidates (h$_y$) \cite{yang2018learning}. The learning rate is set to $8 \times 10^{-4}$, with an Adamax optimizer. The model is fine-tuned for 25 epochs with a batch size of 128.
    
\end{itemize}
We provide the ``context utterance'' as an input and predict the next most probable ``response utterance'' from the model. The model chooses a response according to a softmax on the dot product (h$_x$·h$_y$) out of all candidates. We minimise the negative log-likelihood of selecting the correct response. The utterances from the SE dataset were split into three parts: training data (80\%), validation data (10\%) and test data (10\%). The number of training epochs was decided to avoid over-fitting on the data and due to resource constraints.

%  \subsection{Transformer Model}
% To generate empathetic responses we experimented with a transformer network \cite{Vaswani2017} as it has proven successful in dialogue generation \cite{Zhang2018}. In this method, the transformer model is given a set of candidates $Y$. The candidate set $Y$ consists of all the responses in the batch. We will call the context sequence $x$ and the response sequence $y$. The context and set of responses are encoded using a transformer network. Separate encoders are used for context and response encoding. The model is trained to maximise the likelihood of producing the correct target response given context(x). From the candidate set, the model chooses the "\textit{best}" candidate, based on the softmax of the dot product of the encoded context and candidates. 
% \subsection{BERT-based model}
% We also experimented with the pretrained BERT based model for empathetic response generation as it has been successful for the next sentence prediction. We use the \textit{'bert-base-uncased'} pretrained model by HuggingFace \cite{Wolf2019HuggingFacesTS}.

% This BERT model is fine-tuned keeping the train-validation-test data and their split exactly same as that used for the transformer model. The model is fine-tuned for 12 epochs with a batch size of 16 using embedding dimension of 300, with a learning rate of $5 \times 10^{-5}$, and the Adamax optimizer.

\subsection{Incorporating Emoji}

\begin{table}[t]
\centering
\resizebox{\hsize}{!}{
\includegraphics{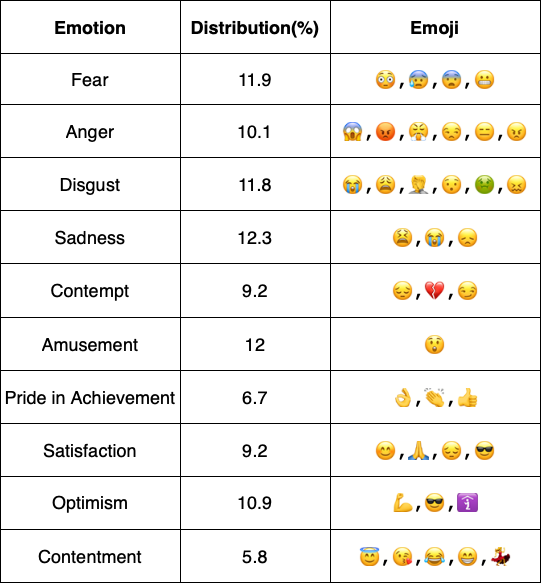}
}
\caption{Distribution of conversations in each emotion and the group of emojis relevant to an emotion}
\label{table2}
\vspace{-0.5cm}
\end{table}

Once we have a text-based response, we append the relevant emoji at the end. We achieve this task by identifying the emotion of the generated response from language models using CNN-based classifier and then selecting the most relevant emoji based on the emotion as shown in Table \ref{table2}.

\begin{itemize}[noitemsep,nosep]
    \item \textbf{Identifying emotion}: Figure~\ref{fig:fig5} shows the architecture of the CNN-based emotion classifier inspired from \citet{Kim2014}. We trained the emotion classifier on the ``Context'' of each conversation as an input and their corresponding ``Emotion'' labels in the SE dataset as an output. We chose ``Context'' attribute of each conversation instead of the utterances because ``Context'' summarises the content of the conversation without directly revealing the details of the conversation. Figure \ref{fig2} shows an example of context and emotion pair. We split the dataset into 72-8-20 for train-validation-test split required for the evaluation and tuning. We trained the model with an Adam optimizer at a learning rate of 0.001, and a decay of $10^{-6}$ for two epochs with a batch size of 128 using cross-entropy loss. After training, we used the emotion classifier with the generated text from language models to obtain the appropriate emotion related to the sentence.
    
    % \todo{- Context because of summary, - Text as input}
    % \todo{Why have you not taken ``context utterance'' and ``emotion'' as input-output pairs for emotion classifier?}

    \item \textbf{Getting relevant emoji}: After getting the generated sentence's emotion, we need a relevant emoji which can be embedded in the text. Using the emotion from the classifier, we obtain a group of emojis which signify the output emotion. We obtain this bucket of emojis using Table \ref{table2}. Table \ref{table2} is obtained by mapping the most commonly used emojis to their corresponding emotion \cite{Novak2015}. After obtaining the bucket, the next step is to get the most relevant emoji from the bucket since the bucket may contain more than one emojis per emotion. To select the most relevant emoji, we compare the cosine similarity between each emoji's embedding and sentence embedding of the generated response.
    % We mapped the most commonly used emojis \cite{Novak2015} to each emotion category (shown in Table \ref{table2}) and selected the most relevant emoji from a given emotion as our final output. 
    
    We obtain the emoji's embedding using Emoji2Vec \cite{Eisner2016} and the word embeddings for the sentence embedding using pre-trained Word2Vec \cite{Demeester2016}. Sentence embedding is generated using the method proposed by \citet{arora2016simple}. Since Emoji2Vec generates embeddings using a pre-trained model of Word2Vec on the words associated with the emoji, we chose to use Word2Vec embeddings for the generated textual response instead of BERT embeddings. This technique helps in providing the same space to sentence and emoji embedding. Finally, the emoji with maximum cosine similarity with sentence embedding is taken as the most relevant emoji from the bucket. We add the emoji at the end of the sentence to generate an empathetic response.
    
    Although, the emotion classifier provide us the emotion imbibed in the generated sentence, still the emotion may not be explicit enough to add an emoji. Thus, only when the cosine similarity is above a threshold, the emoji is added. This way, we avoided adding emojis to all sentences, and hence avoided their unrealistic and excessive use.
    
\end{itemize}

\section{Evaluation}
\begin{table}[t]

\centering
\resizebox{\hsize}{!}{
\smallskip\begin{tabular}{ | l | c | c | }
    \hline
    \multirow{2}{*}{\textbf{Model}} & \textbf{Average} & \multirow{2}{*}{\textbf{P@1,100}} \\
    & \textbf{BLUE Score} & \\
    \hline \hline
    Transformer & 4.38 & 3.65\% \\\hline
    BERT & \textbf{ 5.78 }& \textbf{36\%}\\
    \hline
\end{tabular}
}
\caption{Automatic evaluation metrics on the test set}\smallskip
\label{table3}
\end{table}

\textbf{Automated Metrics:} Following the practice of earlier works in dialogue generation \cite{DBLP:journals/corr/LiGBGD15, wen-etal-2015-semantically}, we compared the model generated response with the actual response using the BLEU scores. The BLEU scores (average of BLEU-1, BLEU-2, BLEU-3, and BLEU-4) of all the samples in the test set were averaged for Transformer and BERT based models. Then, we computed the P@1,100 \cite{Rashkin2018} to evaluate the performance of the response-retrieval systems. Table \ref{table3} summarises the results and shows that BERT-model outperforms the Transformer-based approach in terms of both the metrics.

On evaluating the emotion classifier, we achieved the micro accuracy of 55.4\%, macro accuracy of 54.6\%, and macro F1-score of 55.9\%. According to \citet{Liu2018}, extracting emotions is the biggest challenge in identifying the emoji. Hence, our results are consistent with the experiments by \citet{Liu2018}. Even though the results can be improved with advanced models, our pipeline is an attempt to formalise the problem statement and provide its significance.\\

\noindent \textbf{Human Evaluation:} We evaluate 80 dialogues generated from BERT-based SentEmojiBot: 40 dialogues with emojis and the same 40 dialogues without emojis.  We split the dialogues into four sets of 20 randomly chosen dialogues. All the sets are mutually exclusive from each other. \iffalse 2) none of the subjects get the same dialogue differing only in terms of the presence of emoji .\fi Each set was shared with five English-speaking human evaluators (different from the authors of paper), that evaluated each dialogue on a Likert scale (1--5) \cite{joshi2015likert}. The total number of evaluators were 20. The evaluators rated the dialogues on the basis of two criteria i.e. the empathy of generated dialogue and the relevance of the added emoji. For dialogues without emoji, the relevance of added emoji is not rated. All the ratings are averaged across each of the tasks to obtain the final evaluation score shown in Table \ref{table4}. We observed that emojis improved the empathy score by 0.49. Furthermore, the relevance score of 3.11 reflects that the evaluators feel that the emojis were relevant to the context on the Likert scale.
\begin{table}[t]

\centering
% \resizebox{\hsize}{!}{
\begin{tabular}{ | l | c | c | }
    \hline
    \multirow{2}{*}{\textbf{User-Study}} & \multirow{2}{*}{\textbf{Empathy}} & \multirow{2}{*}{\textbf{Relevance}} \\
    & & \textbf{of emoji} \\ \hline \hline
    Responses & \multirow{2}{*}{2.88/5} & \multirow{2}{*}{-} \\
    without emojis & & \\ \hline
    Responses & \multirow{2}{*}{\textbf{3.37/5}} & \multirow{2}{*}{\textbf{3.11/5}}\\
    with emojis & & \\
    \hline
\end{tabular}
% }
\caption{Human ratings: Empathy and Relevance}
\label{table4}
\end{table}

\section{Discussions And Conclusion} 
We showed the efficacy of emojis to improve empathetic responses and developed a system- SentEmojiBot to generate empathetic responses inculcating emojis. As shown in Table~\ref{table3}, SentEmojiBot performed well in terms of the metrics. The human ratings in Table~\ref{table4} show that added emojis were satisfactory relevant and increased empathy of responses. We hope our pipeline and results will promote more research on using cross-modality data like emojis for improving empathetic behaviour of dialogue agents. Our current work is limited to including emojis (a) at the end of sentences, and (b) after generating text-based dialogues. However, humans often use emojis in between dialogues, hence, in the future, generating emojis as a part of the dialogue itself can be another direction to make the response more natural and empathetic.

\bibliography{anthology,eacl2021}
\bibliographystyle{acl_natbib}

\end{document}